\documentclass{article}
\usepackage{subcaption}
\usepackage{microtype}
\usepackage{graphicx}
\usepackage{booktabs} 
\usepackage{custom_style}

\usepackage{hyperref}
\usepackage{multirow}
\usepackage{mdframed}
\usepackage{graphicx}
\usepackage{svg}

\usepackage[preprint]{icml2025}

\usepackage{amsmath}
\usepackage{amssymb}
\usepackage{mathtools}
\usepackage{enumitem}
\usepackage{graphicx} 
\usepackage{amsthm}

\usepackage[capitalize,noabbrev]{cleveref}

\theoremstyle{plain}
\newtheorem{theorem}{Theorem}[section]

\theoremstyle{definition}
\newtheorem{definition}[theorem]{Definition}

\theoremstyle{remark}

\usepackage[disable,textsize=tiny]{todonotes}

\icmltitlerunning{Toward a Flexible Framework for Linear Representation Hypothesis}

\begin{document}

\twocolumn[
\icmltitle{
Toward a Flexible Framework for Linear Representation Hypothesis \\ Using  Maximum Likelihood Estimation }



\icmlsetsymbol{equal}{*}

\begin{icmlauthorlist}
\icmlauthor{Trung Nguyen}{cs}
\icmlauthor{Yan Leng}{mc}
\end{icmlauthorlist}

\icmlaffiliation{cs}{Department of Computer Science, University of Texas at Austin, TX, USA}
\icmlaffiliation{mc}{McCombs School of Business, University of Texas at Austin, TX, USA}

\icmlcorrespondingauthor{Trung Nguyen}{trungnguyen@utexas.edu}

\icmlkeywords{Interpretability, Large Language Models, Linear Representation Hypothesis, Activation Engineering, Maximum Likelihood Estimation}
\vskip 0.3in
]



\printAffiliationsAndNotice{}  

\begin{abstract}
Linear representation hypothesis posits that high-level concepts are encoded as linear directions in the representation spaces of LLMs. \citet{park2024the} formalize this notion by unifying multiple interpretations of linear representation, such as 1-dimensional subspace representation and interventions, using a causal inner product.  However, their framework relies on single-token counterfactual pairs and cannot handle ambiguous contrasting pairs, limiting its applicability to complex or context-dependent concepts. 
We introduce a new notion of binary concepts as unit vectors in a canonical representation space, and utilize LLMs' (neural) activation differences along with maximum likelihood estimation (MLE) to compute concept directions (i.e., steering vectors). 
Our method, Sum of Activation-base Normalized Difference (\textsf{SAND}), formalizes the use of activation differences modeled as samples from a von Mises-Fisher (vMF) distribution, providing a principled approach to derive concept directions. 
We extend the applicability of \citet{park2024the} by eliminating the dependency on unembedding representations and single-token pairs. 
Through experiments with LLaMA models across diverse concepts and benchmarks, we demonstrate that our lightweight approach offers greater flexibility, superior performance in activation engineering tasks like monitoring and manipulation.
\end{abstract}

\section{Introduction}
The linear representation hypothesis (LRH) posits that high-level concepts are encoded as linear directions in a representation space, providing a structured framework for understanding how concepts are embedded and manipulated in large language models\footnote{In this work, the term ``Large Language Model (LLM)" refers specifically to decoder-only, autoregressive models designed for text generation.} (LLMs)~\citep{singhrepresentation, jiangorigins}. 
This hypothesis implicitly forms the theoretical foundation for many studies in the emerging field of representation engineering (also known as activation engineering), which focuses on designing, transforming, and manipulating LLM representations for applications such as probing, steering, and concept erasure. 
While strong empirical evidence supports the connection between LRH and representation engineering~\citep{zou2023transparency, rimsky-etal-2024-steering, li2024inference}, their theoretical relationship remains less well understood.
\citet{park2024the} take an important step in this direction by unifying three interpretations of linear representations through a causal inner product, which maps unembedding representations to embedding representations.

Despite the significance of \citet{park2024the}, it has several limitations. It restricts binary concepts to single-token counterfactual pairs, making it unsuitable for more complex, context-dependent concepts such as “untruthful\( \to \)truthful,” which cannot be adequately represented by individual tokens. Furthermore, token-based representations are often ambiguous, as a single token pair can correspond to multiple overlapping or unrelated concepts. For example, the pair (“king”, “queen”) may represent “male\(\to\)female,” “k-words\(\to\)q-words,” or “n-th card \(\to\) (n-1)-th card,” depending on the context. Additionally, the reliance on unembedding representations and causal inner products limits the flexibility of representation construction. 

This work bridges the gap between the theory of the linear representation hypothesis and the practice of representation engineering by tackling two key limitations.
First, prior studies rely on restrictive definitions of binary concepts, which limit their applicability to more general concepts.  
Second, they require single-token counterfactual pairs to distinguish concepts, which introduces inconsistencies and fails to account for the broader context of language models. To overcome these limitations, we introduce a generalized framework that redefines representations in a canonical representation space, inspired by the unified representation proposed by \citet{park2024the}.

Building on the intuition that activation differences between positive and negative prompts (e.g., “truthful” vs. “untruthful”) capture the direction of a concept in the model’s activation space, we propose a method that formalizes and generalizes this idea. 
Specifically, we assume a canonical representation space obtained from the LLM activation space via a mapping $\Psi$, such that activation differences are mapped to samples from a von Mises-Fisher (vMF) distribution whose mean direction representing the binary concept. 
Using MLE, we derive an estimator for the concept direction in the canonical space and map it back to the activation space via a transformation $\Psi^{-1}$. 
This results in a simple yet effective method, which we term \underline{S}um of \underline{A}ctivation-base \underline{N}ormalized \underline{D}ifferences (SAND), for computing concept directions. 

Our framework avoids reliance on restrictive definitions such as binary concepts or single-token counterfactual pairs, offering a lightweight and generalizable approach for representation engineering. This method has broad applicability, enabling probing and steering in LLMs.

We bridge the gap between the linear representation hypothesis and representation engineering, offering a unified framework for probing and manipulating LLMs. Our approach not only enhances theoretical understanding but also provides practical tools for real-world applications in concept control and LLM interpretability.

Our work makes the following contributions to the linear representation and representation engineering literature: 
\begin{itemize}
    \item For the linear representation literature, we introduce a new framework that redefines binary concepts as unit vectors in a canonical representation space, addressing limitations in prior methods that rely on single-token counterfactual pairs and ambiguous token-based representations (Section~\ref{subsec:framework}). 
    \item We propose a novel method to construct concept directions using activation differences, formalized through a von Mises-Fisher (vMF) distribution and maximum likelihood estimation (MLE), offering a principled and robust approach for the growing representation engineering literature (Section~\ref{subsec:algo}). 
    \item We provide theoretical insights into the empirical effectiveness of the heuristic Mean Difference method for extracting steering vectors (Section~\ref{subsec:sand_md_connect}).
    \item Our method yields Algorithm~\ref{al:main} that can be incorporated into state-of-the-art activation engineering frameworks at a minor computational cost of one matrix multiplication (Sections~\ref{subsec:sand},~\ref{subsec:op_cnt}).
    \item We validate the proposed framework through extensive experiments with LLaMA models, demonstrating its effectiveness in constructing concept directions and advancing practical and lightweight tools for representation engineering (Section~\ref{sec:expe}).  
\end{itemize}

\section{Related Work} 
\paragraph{Linear Representation Hypothesis}
The linear representation hypothesis suggests that human-interpretable concepts are encoded as linear directions or subspaces within an LLM’s representation space. This implies that LLM behavior can be understood and controlled by steering residual stream activations along these directions \citep{singhrepresentation, zou2023transparency}.

\citet{park2024the} unified these notions of linear representation under the framework of a causal inner product, providing theoretical foundations for the hypothesis. 

\citet{jiangorigins} investigated the origins of linear representations by introducing a latent variable model where context sentences and next tokens share a latent space. They proved that latent concepts emerge as linear structures within the learned representation space. 

Numerous studies provide empirical evidence that high-level concepts—including political ideology, sentiment \citep{tigges2023linear, hollinsworth2024language}, truthfulness \citep{zou2023transparency, li2024inference, marks2023geometry}, humor \citep{vonlanguage}, safety \citep{arditi2024refusal}, and even abstract notions like time and space \citep{gurnee2023language}—are linearly encoded in LLM representations. This growing body of work underscores the significant potential of linear representation for interpreting and influencing model behavior. 

Our study proposes a new framework that redefines binary concepts as unit vectors in a canonical representation space. This framework overcomes the limitations of prior methods that depend on single-token counterfactual pairs and ambiguous token-based representations, allowing for more general and context-aware representation engineering.

\paragraph{Concept Vector for Activation Engineering}
\label{subsec:steering_activation_engineering}

Steering vectors, used in activation engineering to control LLMs at inference time \citep{li2024inference, zhao2024steering}, can be categorized into four groups: activation-difference, linear probing, unsupervised, and training-based methods.

Activation-difference methods, the most widely used approach, compute steering vectors by leveraging differences in activations from contrasting prompts. \textit{Activation Addition (ActAdd)} derives vectors from a single prompt pair \citep{turner2024steeringlanguagemodelsactivation}, while \textit{Contrastive Activation Addition (CAA)} extends this to datasets of contrasting pairs for greater robustness \citep{rimsky-etal-2024-steering}. Variants include deriving vectors from activation differences between target and misaligned teacher models \citep{wang2024trojan} or mitigating biases through contrastive differences \citep{chu2024causal, arditi2024refusal}. Techniques like mean-centering refine these vectors by aligning them with dataset-specific properties \citep{jorgensen2023improving, postmus2024steering, panickssery2023steering}. \citet{singhrepresentation} provide theoretical justification for mean-difference steering, showing that simple additive steering is optimal under certain constraints. 

Linear probing methods use probe weight directions derived from supervised method, such as regression and linear discriminant analysis, trained to distinguish between contrasting datasets~\citep{zhao2024steering, mallen2023eliciting, park2024the}. However, they perform significantly worse than activation-difference approaches in a truthfulness steering application~\citep{li2024inference}. 

Unsupervised dimensionality reduction methods, such as Principal Component Analysis (PCA), identify important directions in activation space or reduce dimensionality before deriving steering vectors \citep{zou2023transparency, liu2023context, adiladiscovering, wu2024reft, park2024the, burnsdiscovering}. These techniques effectively isolate concept-specific directions, such as biases or stylistic features.

Training-based methods include latent steering vectors, derived through gradient descent for target-specific outputs \citep{subramani2022extracting}, and bi-directional preference optimization, which optimizes vectors using contrastive human preferences \citep{cao2024personalized}. Conceptor methods use soft projection matrices to represent activation covariance \citep{postmus2024steering}, while sparse autoencoders extract interpretable features from activations for steering \citep{o2024steering, zhao2024steering}. These methods are precise but computationally intensive due to iterative optimization and high resource demands.


Our study introduces a novel method for constructing steering vectors by integrating vMF distributions with MLE. This approach is low-cost, robust, and principled. These properties enable flexible and effective applications such as concept probing and directional manipulation in LLMs. 

\section{Background: Revisiting \citet{park2024the}}
\label{sec:lin_rep}
\comment{This work uses some materials and address some shortcomings of \cite{park2024the}.}

We first review the framework proposed by \citet{park2024the}, which motivates our work.
\citet{park2024the} models the probabilities distribution over next tokens as
\[
\Pr[y|x] \propto \exp(\lambda(x)^T\gamma(y))
\]
where $\lambda(x)$ is the context embedding of an input $x$ (i.e., the output embedding for the last token from the last transformer layer) and $\gamma(y)$ is the unembedding of a token $y$. 

\paragraph{Binary Concepts and Causal Separability.}
To formalize binary concepts, 
\citet{park2024the} introduce a latent variable $W$ that is caused by the context $X$ and generates the output $Y$ such that $Y(W=w)$ only depends on $w \in\{0, 1\}$. 
Two concepts $W, Z$ are called \textit{causally separable} if $Y(W=w, Z=z)$ is well-defined for each $w, z$. 

\citet{park2024the} then define an \textit{unembedding representation} $\overline{\gamma}_W$ of a concept $W$ if $\gamma(Y(1)) - \gamma(Y(0)) = \alpha \overline{\gamma}_W$ for some $\alpha>0$ almost surely.

There are two limitations of these definition. First, their method was restricted to work on only binary concepts that can be differentiated by single-token counterfactual pairs of outputs, such as ``male$\to$female", ``English$\to$French"~\citep{anonymous2025intricaciesoffeature, park2024geometrycategority}. 
This means that the approach is limited in its ability to capture complex, real-world concepts that do not have a clear binary opposition or a single token that indicates their presence or absence.  
For example, concepts like \emph{truthfulness} do not map to specific token pairs. The statement ``The earth is flat'' is untrue, but one cannot identify a single token that makes it \emph{untruthful}. 
In general, a concept can be expressed across a phrase, sentence, or paragraph, and is not always reducible to a single token or a pair of tokens.

Second, each pair of counterfactual tokens $(Y(0), Y(1))$ can in fact corresponds to multiple different concepts. 
For instance, (``king'', ``queen'') can represent ``female$\to$male", ``k-words$\to$q-words", and "n-th card$\to$(n-1)-th card" in a deck of playing cards. In general, tokens and words, when presented alone, are frequently ambiguous and can have multiple potential meanings or interpretations. This ambiguity makes it challenging to isolate the specific concept of interest using only counterfactual pairs.

\paragraph{Linear Representation in the Embedding Space.}
\citet{park2024the} define a notion of linear representation in the embedding space as follows.
\begin{definition}
\label{def:park_emb}
$\overline{\lambda}_W$ is an embedding representation of a concept $W$ if we have
$\lambda_1 - \lambda_0\in Cone(\overline{\lambda}_W)$ for any context embeddings $\lambda_0, \lambda_1$ that satisfy
\[
\frac{\Pr[W=1|\lambda_1]}{\Pr[W=1|\lambda_0]} > 1
\]
and \[
\frac{\Pr[W, Z|\lambda_1]
}{
\Pr[W, Z|\lambda_0]
} = \frac{\Pr[W|\lambda_1]}{\Pr[W|\lambda_0]}
\] for each concept $Z$ that is causally separable with $W$.    
\end{definition}

Intuitively, adding $\overline{\lambda}_W$ to an embedding $\lambda_0$ steers the model toward outputs consistent with $W=1$ without affecting outputs for concepts that are causally separable from $W$.

\paragraph{Unified Representations via the Causal Inner Product.}
Next, \citet{park2024the} introduce a 
causal inner product $\langle\cdot, \cdot\rangle_C$ on the unembedding space. 
For any pairs of causally separable concepts $W$ and $Z$, their unembedding representations satisfy 
$\langle\overline{\gamma}_W, \overline{\gamma}_Z\rangle_C = 0$. 

\citet{park2024the} show that the Riesz isomorphism with respect to a causal inner product maps unembedding representations to their embedding counterparts, enabling them to leverage the former for constructing the latter. 

Finally, a concrete example of a causal inner product from \citet{park2024the} is 
\[
\langle\overline{\gamma}, \overline{\gamma}'\rangle_C := \overline{\gamma}^T Cov(\gamma)^{-1} \overline{\gamma}'
\] where $\gamma$ is the unembedding vector of a token sampled uniformly at random from the vocabulary. 
It leads to the following unified representations for each concept $W$, $\bar{g}_W=\bar{l}_W$ where $\bar{g}_W:=Cov(\gamma)^{-1/2} \overline{\gamma}_W$ and $\bar{l}_W:=Cov(\gamma)^{1/2} \overline{\lambda}_W$.

\section{Our Proposed Framework}
\label{sec:framework}
In this section, we present our framework and introduce our algorithm, along with its computational complexity. 

\subsection{Preliminaries} 
A von Mises-Fisher (vMF) distribution on the unit sphere \( \mathbb{S}^p \) is parameterized by a mean direction \( \mu \) and a concentration parameter \( \kappa \), with the density function:
\[
f(x|\mu, \kappa) = c_p(\kappa) e^{\kappa \mu^T x},
\]
where \( x \in \mathbb{S}^p \) is a unit vector, \( \mu \in \mathbb{S}^p \) is the mean direction, and \( c_p(\kappa) \) is a normalization constant~\cite{sra2012short}. The vMF distribution is among the simplest models for directional data, and mirrors many properties of the multivariate Gaussian distribution in \(\bbR^d\). 

\subsection{Generalized Representation Framework}
\label{subsec:framework}
To sum up Section~\ref{sec:lin_rep}, three definitions--binary concepts, causal separability and unembedding representations--all formulated around single-token counterfactual pairs. 
These definitions can be impractical in real-world scenarios (e.g., where a concept like “truthfulness” cannot be captured by a token-level change). 
However, these three definitions are used to construct embedding representations (i.e., concept directions) that can steer model outputs toward (or away from) a target concept.

In this work, we remove these restrictive definitions while preserving the ability to obtain effective concept directions for monitoring and manipulating LLM internals.
We start by assuming an imaginary canonical representation space, implicitly corresponding to the unified representation space in \citet{park2024the}, and treat {each binary concept as a unit vector} therein:
\begin{definition}
    A binary concept is a unit vector in this canonical space.
\end{definition} 
To relate this mathematical definition to the human natural language understanding of a binary concept such as ``untruthful $\to$ truthful", we use LLMs' activation spaces as bridges. Precisely, we use a map $\Psi$ to map LLM activations to representations in the canonical space and a map $\Psi^{-1}$ to map in the opposite direction. Although the mappings $\Psi$ and $\Psi^{-1}$ can be linear or non-linear, and layer-dependent, since our canonical space is implicitly referred to the unified space in \citet{park2024the}, we examine two linear choices of $\Psi$ explored in their work.
\begin{enumerate}[label=(\roman*)]
\item \textbf{Identity map.} $\Psi$ is the identity, so that the canonical space and the activation space coincide.
\item \textbf{Whitening map.} $\Psi = \mathrm{Cov}(\gamma)^{1/2}$, following the causal-inner-product example in \citet{park2024the}.
\end{enumerate}

Rather than using unembedding representations and Riesz isomorphisms, we use (neural) activation differences 
\citep{zou2023transparency, turner2024steeringlanguagemodelsactivation} along with maximum likelihood estimation (MLE) to construct concept directions. These concept directions are also termed ``reading vectors" or ``embedding representations" in the literature.

\subsection{Deriving Our Algorithm}
\label{subsec:algo}
We formalize the estimation of concept directions using activation differences, vMF distributions, and MLE.   
Let $\bar{l}$ be a binary concept in the canonical space. Thus, its image in the LLM activation space is given by
\[
\overline{\lambda} = \Psi^{-1}\bar{l}.
\]

We call $\overline{\lambda}$ a concept direction in the activation space.\\
To estimate $\bar{\lambda}$ from data, we leverage activation-difference methods (see review in Section~\ref{subsec:steering_activation_engineering}). 
Concretely, we select contrasting pairs of prompts $\{p_i^+, p_i^-\}$, where $p^+_i$ represents the desired property or concept (e.g., ``love") and $p^-$ is an opposing or neutral counterpart (e.g., containing \emph{hate}).
Let $h^+_l$ be the activation vector for the positive prompt \(p^+\) at layer \(l\). Let $h^-_l$ be the activation vector for the negative prompt \(p^-\) at layer \(l\). 
The difference vector \(h^+_l - h^-_l\) is viewed as a direction capturing how the model’s internal representation shifts when switching from a negative to a positive instance of the concept. 

In the following, we denote activation differences as a set of vectors $\Lambda = \{\Tilde{\lambda}_1, \Tilde{\lambda}_2, \ldots, \Tilde{\lambda}_k\}$ in the activation space.\footnote{The model's activation space is sometimes referred to as the context embedding space in the literature. }
We formalize the intuition that activation differences capture the essence of the concept direction as follows: Set $\{\Tilde{l}_1, \ldots, \Tilde{l}_k\}$ follow a vMF distribution whose mean is $\bar{l}$, where $\Tilde{l}_i := \frac{\Psi \Tilde{\lambda}_i}{\norm{\Psi \Tilde{\lambda}_i}}$ and $\norm{\cdot}$ refers to the 2-norm of vectors or matrices. The MLE for \( \bar{l} \) is then given by:
\[
\hat{\bar{l}} = \frac{\sum_{i=1}^k \Tilde{l}_i}{\|\sum_{i=1}^k \Tilde{l}_i\|} \uparrow \sum_{i=1}^k \Tilde{l}_i.  
\]
Here, given two vectors $v_1, v_2$, we say $v_1$ and $v_2$ point in the same direction, denoted as $v_1\uparrow v_2$ if there exists a positive number $c$ such that $v_1 = c \times v_2$.

Using MLE's invariance property \citep[p. 320]{casella2002}, the MLE for \(\bar{\lambda}\) is given by
\begin{equation}
\label{eq:main}
\hat{\bar{\lambda}} = \Psi^{-1}\hat{\bar{l}} \uparrow \sum_{i=1}^k \frac{\Tilde{\lambda}_i}{\norm{\Psi\Tilde{\lambda}_i}}.    
\end{equation}
One can interpret Equation~\ref{eq:main} as the sum of normalized activation differences (with respect to $\Psi$). Thus, we term this method "\underline{S}um of \underline{A}ctivation-based \underline{N}ormalized \underline{D}ifferences", or \textsf{SAND} for short.
\subsection{Choices for Geometry in Activation Space $\Psi$}
\label{subsec:psi_choices}
One can also interpret Equation~\ref{eq:main} as using \(\Psi\) to define a new norm on the activation space, thereby shaping its geometry.\\
In this work, we experiment with two choices of $\Psi$. The first choice is the simple identity matrix. 
This map implies that the canonical and activation spaces coincide, and it reduces Equation~\eqref{eq:main} to
\begin{equation}
\label{eq:eucl}
\hat{\bar{\lambda}}\uparrow \sum_{i=1}^k \frac{\Tilde{\lambda}_i}{\norm{\Tilde{\lambda}_i}}.    
\end{equation}
In other words, the $\Psi$-norm is just the usual Euclidean norm, so we take the sum of each activation-difference vector normalized by its length.

The second choice is the whitening transformation used in the causal-inner-product approach of \citet{park2024the}. Let $E \in \mathbb{R}^{n_v \times d}$ be the embedding matrix of an LLM, which has a row for each of $n_v$ tokens in the vocabulary. Consider picking uniformly at random a row $\gamma$ of $E$. Let $\bar{\gamma} = \mathbb{E}[\gamma]$, and $C$ be the matrix obtained by subtracting $\bar{\gamma}$ from each row of $E$. Thus, the covariance matrix of $\gamma$ is given by 
\[
Cov(\gamma) = \frac{C^TC}{n_v}.
\]
Let $\Psi:=Cov(\gamma)^{1/2}$. 

Some simple algebra gives 
\[
\norm{\Psi \Tilde{\lambda}_i} = \sqrt{ \Tilde{\lambda}_i^T Cov(\gamma)\Tilde{\lambda}_i} = \sqrt{ \Tilde{\lambda}_i^T \frac{C^TC}{n_v}\Tilde{\lambda}_i} = n_v^{-1/2}\norm{C\Tilde{\lambda}_i}.
\]
Hence, 
\begin{equation}
\label{eq:whit}
\hat{\bar{\lambda}}\uparrow \sum_{i=1}^k \frac{\Tilde{\lambda}_i}{\norm{C\Tilde{\lambda}_i}}.
\end{equation}

\subsection{The \textsf{SAND} Algorithm}
\label{subsec:sand}
To efficiently implement the sums in Equations~\eqref{eq:eucl} and \eqref{eq:whit}, we collect all activation-difference vectors $\tilde{\lambda}_i$  as columns of a matrix $\Lambda \in \mathbb{R}^{d \times k}$. Likewise, let $C \in \mathbb{R}^{n_v \times d}$ be the mean-subtracted embedding matrix described in Section~\ref{subsec:psi_choices}. Given these matrices, Equations~\eqref{eq:eucl} and \eqref{eq:whit} yield Algorithm~\ref{al:main} (\textbf{\textsf{SAND}}: \underline{S}um of \underline{A}ctivation-based \underline{N}ormalized \underline{D}ifferences). This procedure can be fully vectorized and is readily implemented on modern hardware via state-of-the-art software packages such as NumPy, SciPy, PyTorch, TensorFlow, or MATLAB.\\
In Algorithm~\ref{al:main}, $\odot, \oslash, \sqrt{\cdot}$ denote element-wise multiplication, division, and square root respectively, and \(\text{sum}(\cdot, \, \text{axis}=0)\) refers to column-wise summation of matrix entries. 

\begin{algorithm}[h!]
\caption{SAND: \underline{S}um of \underline{A}ctivation-based \underline{N}ormalized \underline{D}ifferences}
\begin{algorithmic}[1]
\footnotesize 
\label{al:main}
\REQUIRE Matrix $\Lambda \in \mathbb{R}^{d \times k}$ with columns $\Tilde{\lambda}_i$ for $i = 1, \dots, k$, matrix $C \in \mathbb{R}^{n_v \times d}$
\ENSURE $S_1 = \sum_{i=1}^k \frac{\Tilde{\lambda}_i}{\|\Tilde{\lambda}_i\|}$ and $S_2 = \sum_{i=1}^k \frac{\Tilde{\lambda}_i}{\|C\Tilde{\lambda}_i\|}$

\STATE \textbf{Step 1: Compute column-wise norms of $\Lambda$}:
\[
\mathbf{N}_1 \gets \sqrt{\text{sum}(\Lambda \odot \Lambda, \, \text{axis}=0)}
\]
\STATE \textbf{Step 2: Compute transformed matrix $C\Lambda$}:
\[
\Lambda_C \gets C \cdot \Lambda
\]
\STATE \textbf{Step 3: Compute column-wise norms of $C\Lambda$}:
\[
\mathbf{N}_2 \gets \sqrt{\text{sum}(\Lambda_C \odot \Lambda_C, \, \text{axis}=0)}
\]
\STATE \textbf{Step 4: Compute normalized sums}:
\[
S_1 \gets \Lambda \cdot \left(\mathbf{1}_k \oslash \mathbf{N}_1\right)
\]
\[
S_2 \gets \Lambda \cdot \left(\mathbf{1}_k \oslash \mathbf{N}_2\right)
\]

\STATE Output $S_1$ and $S_2$
\end{algorithmic}
\end{algorithm}

\subsection{Operation Count}
\label{subsec:op_cnt}
To formally assess the computational cost of Algorithm~\ref{al:main}, we follow the classical approach and count the number of floating point operations (flops), where each addition, subtraction, multiplication, division, or square root counts as one flop~\citep[p. 59]{doi:10.1137/1.9781611977165}.
\begin{theorem}
\label{theo:cost}
    Algorithm~\ref{al:main} requires \(\sim 2n_v\times d\times k \) flops given input matrices $\Lambda \in \mathbb{R}^{d \times k}$ and $C \in \mathbb{R}^{n_v \times d}$,
\end{theorem}
where the symbol ``$\sim$" means
\[
\lim_{d, k, n_v\to\infty}\frac{\text{number of flops}}{2n_v\times d\times k} \leq 1.
\]
Theorem~\ref{theo:cost} can be established as follows. Step 1 requires $d\times k$ multiplications, followed by $(d-1)\times k$ additions, and $k$ square roots. Thus, in total, Step 1 requires \(2d\times k\) flops. For matrix multiplication in Step 2, the straightforward computation requires $\sim 2d\times n_v\times k$ flops. Step 3 is counted similarly to step 1, and requires $2n_v \times k$ flops. Finally, Step 4 requires $2k$ divisions, followed by two matrix-vector multiplications, each requires $(2d-1)\times k$ flops. Thus the total flop count for Step 4 is \(2d\times 2k\). 
Therefore, the total cost of Algorithm~\ref{al:main} is dominated by the matrix multiplication in Step 2, and is $\sim 2n_v\times d\times k$ flops.


Because the major expense is a single $\mathbf{(n_v \times d)\cdot(d \times k)}$ multiplication, \textsf{SAND} can be incorporated into any existing activation-engineering pipelines (see papers reviewed in Section~\ref{subsec:steering_activation_engineering}) at essentially the cost of one matrix multiplication, which is usually negligible compared to large-scale inference or training.

\section{Experiments}
\label{sec:expe}

We first explore the relationship between \textsf{SAND}, with different geometric choices in the activation spaces, and the widely-used heuristic method, Mean Difference. We further investigate why different choices in $\Psi$ lead to similar concept directions by analyzing the spectrum of matrices $C$.

We then explain why \textsf{SAND} can identify the concept direction, aligning with the linear representation hypothesis introduced in Definition~\ref{def:park_emb}. 
Finally, we demonstrate how \textsf{SAND} can be used to monitor the truthfulness of the model. 



\subsection{Connection between \textsf{SAND} (with Different Geometry $\Psi$) and Mean Difference}
\label{subsec:sand_md_connect}
Mean Difference (MD) is a heuristic method used in the literature~\cite{turner2024steeringlanguagemodelsactivation, rimsky-etal-2024-steering, wang2024trojan}, and is the basis for mean-centering approaches  \citep{jorgensen2023improving, postmus2024steering}. 
\citet{zou2023transparency} show that MD achieves top-2 performance in the Correlation task and secures top-1 performance in both the Manipulation and Termination tasks on the Utilitarianism dataset~\citep{hendrycks2021ethics}, where tasks correspond to the concept of utility.

The calculation for MD is similar to Equation~\eqref{eq:eucl}, except for normalization, and can be expressed using our notations:
\begin{equation}
    \hat{\bar{{\lambda}}} \uparrow \sum_{i=1}^k \Tilde{\lambda}_i.
\end{equation}

In this section, we discuss the connection between the high performance of \textsf{SAND} and MD by calculating cosine similarities between concept directions learned by these methods and Principal Component Analysis (PCA) under considered experimental settings. We denote Equation~\eqref{eq:eucl} as SAND-e and Equation~\eqref{eq:whit} as SAND-w. 

We experiment with two concepts: truthfulness and utility. To extract the truthfulness direction, we use six question-answering (QA) examples, each consisting of a question, a correct answer, and an incorrect answer. These examples are provided in Table~\ref{tab:truth_examples} in Appendix~\ref{app:truth}. For utility, we use scenario pairs from the Utilitarianism dataset within the ETHICS benchmark~\cite{hendrycks2021ethics}, where one scenario exhibits higher utility than the other. We vary the number of scenario pairs, using sample sizes of 20, 50, 100, and 1000.

\begin{figure}[h]
    \centering
    \includesvg[width=0.45\textwidth]{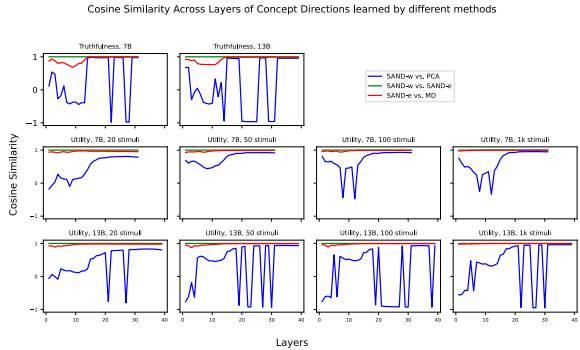}
    \caption{MD, SAND-e, and SAND-w demonstrate significantly stronger alignment in their concept directions compared to PCA. Enlarged versions of these plots are provided in the Appendix~\ref{app:cos}.} 
    \label{fig:cos_sim}
\end{figure}

Figure~\ref{fig:cos_sim} illustrates that MD, SAND-e, and SAND-w exhibit much greater alignment in their concept directions compared to PCA, especially in the middle to final layers, even with as few as six stimuli. 

We hypothesize that SAND-e and MD learn similar embedding representations in our experiments due to the phenomenon of ``anisotropy"~\cite{ait-saada-nadif-2023-anisotropy, godey-etal-2024-anisotropy, machina-mercer-2024-anisotropy, razzhigaev-etal-2024-shape}, wherein transformer embeddings are clustered in a narrow cone.

\begin{figure}[h]
    \centering
    \includesvg[width=0.45\textwidth]{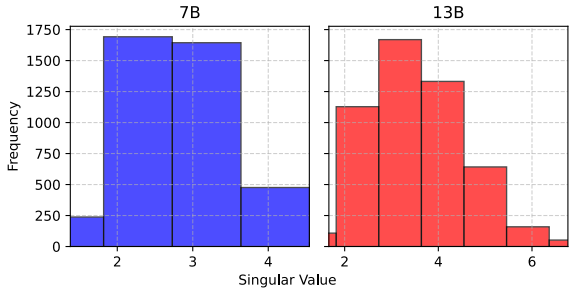}
    \caption{Singular Values within the 1\% to 99\% quantile ranges of Matrices $C$ in LLaMA-2 Chat Models}
    \label{fig:hist_sing_val}
\end{figure}
\begin{figure}[h]
    \centering
    \includesvg[width=0.45\textwidth]{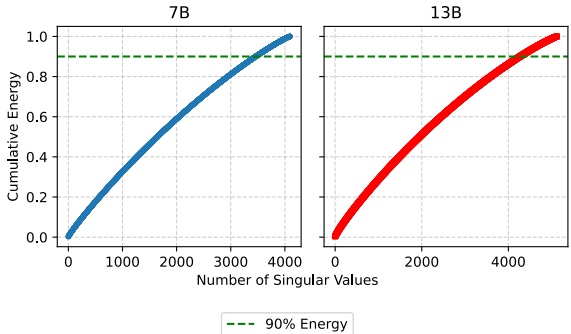}
    \caption{Cumulative Energy Plots of Singular Values for Matrices $C$ in LLaMA-2 Chat Models}
    \label{fig:energy_sing_val}
\end{figure}

\paragraph{Analysis of Spectrum of Matrices \(C\)} 
To understand why SAND-e and SAND-w learn highly similar concept directions in our experiment, we visualize the spectrum of matrices $C$ in Equation~\eqref{eq:whit} for the LLaMA2-7B and 13B Chat models. 
Both models yield well-conditioned matrices $C$. 
Figure~\ref{fig:hist_sing_val} shows singular values are tightly clustered in a narrow range. In addition, Figure~\ref{fig:energy_sing_val} illustrates the cumulative energy curves rise steadily, suggesting that the majority of singular values contribute meaningfully. Consequently, activation differences are stretched at comparable scales under $C$, leading Equations~\eqref{eq:eucl} and~\eqref{eq:whit} to produce similar concept directions.

\subsection{Monitoring Internal Activations}
Monitoring refers to the process of observing and tracking the internal states of LLMs to understand how they are processing information and generating outputs~\citep{zou2023transparency}. 
Monitoring is important because it provides insights on the model's inner workings, identify potential issues, and ensure that the model behaves in a safe, ethical, and reliable manner~\citep{chu2024causal}. 
We evaluate the effectiveness of the concept direction from \textsf{SAND} in monitoring honesty within LLMs' internal states across a variety of QA datasets. 


Linear Artificial Tomography (LAT)~\citep{zou2023transparency} extracts and monitors vector representations of concepts like honesty and utility. It involves designing stimuli, collecting neural activity, and building a linear model to identify patterns. LAT scans can detect deceptive neural activity across model layers. We evaluate \textsf{SAND} by integrating it into LAT for this monitoring task and the next intervention application.



\paragraph{TruthfulQA}
\label{sec:mc1}
The TruthfulQA benchmark evaluates a model's ability to distinguish factual information from a carefully selected set of misleading or incorrect statements. 
Due to the importance of truthfulness of LLMs, this data has been widely studied in the literature~\citep{li2024inference, arditi2024refusal, zou2023transparency}.
The questions are accompanied by false answers designed to be statistically tempting. 
The sub-task MC1 in TruthfulQA is currently the most challenging for LLMs, with the highest reported accuracy of \(59\%\) achieved by GPT-4 (RLHF)~\cite{achiam2023gpt}.
The source for stimuli is the six QA primer examples used in the original zero-shot setup of TruthfulQA, each paired with a corresponding false response generated by LLaMA-2-Chat-13B, which are provided in Table~\ref{tab:truth_examples} in Appendix~\ref{app:truth}. For each trial, we randomize the order of choices in each QA primer~\citep{zou2023transparency}.\footnote{While this randomness has a minor effect on the resulting PCA components, it does not alter the directions computed with \textsf{SAND}, which explains the standard errors of 0.}
Table~\ref{tab:truthfulqa} shows that LAT-\textsf{SAND} consistently outperforms LAT-PCA, as well as zero-shot evaluations using LLaMA-2 or GPT-4. 

\begin{table}[t]
\caption{TruthfulQA MC1 accuracy on three LLaMA-2 Chat models, evaluated using standard (Zero-Shot - S), heuristic (Zero-Shot - H), LAT - PCA, and LAT - \textsf{SAND}. The LAT stimulus set includes six QA primers for both training and validation. Mean accuracy is reported across 15 trials, using the layer selected via the validation set. Parentheses indicate standard errors. Zero-shot and LAT-PCA results are from~\citep[Table 8, Appendix B.1]{zou2023transparency}.
}
\label{tab:truthfulqa}
\vskip 0.15in
\begin{center}
\begin{small}
\begin{sc}
\begin{tabular}{lcc}
    \toprule
    & Zero-Shot & LAT \\
    \cmidrule(r){2-2} \cmidrule(r){3-3}
    & S / H & PCA / \textsf{SAND} \\
    \midrule
    7B  & 31.0 / 32.2 & 58.2 (0.4) / \textbf{59.7} (0.0)\\
    13B & 35.9 / 50.3 & 54.2 (0.2) / \textbf{56.2} (0.0)\\
    70B & 29.9 / 59.2 & 69.8 (0.2) / \textbf{71.1} (0.0)\\
    \midrule
    Average & 32.3 / 47.2 & 60.7 / \textbf{62.3} \\
    \bottomrule
\end{tabular}
\end{sc}
\end{small}
\end{center}
\footnotesize
\textbf{Note:} To ensure a fair comparison, we reproduced results for LAT-PCA in \citet{zou2023transparency} and present them alongside (see Tables~\ref{tab:5benchmark_repr} and~\ref{tab:truthfulqa_repr} in Appendix~\ref{app:truth}). Based on this analysis, we exclude specific (model, benchmark) pairs from our comparison in Tables~\ref{tab:truthfulqa},~\ref{tab:lat-benchmarks} if the originally reported means fall outside the corresponding 95\% confidence intervals. Specifically, we exclude (LLaMA-2 13B Base, RACE) and (LLaMA-2 70B Base, RACE).
\vskip -0.1in
\end{table}
\paragraph{Monitoring Using Other Standard QA Benchmarks} 
To further evaluate the models, we include five additional QA datasets: OpenBookQA~\citep{OpenBookQA2018} for general knowledge and common sense, CommonSenseQA~\citep{talmor-etal-2019-commonsenseqa} for everyday concepts, RACE~\citep{lai-etal-2017-race} for reading comprehension, and ARC~\citep{Clark2018ThinkYH} (which includes both ARC-\underline{E}asy and ARC-\underline{C}hallenge) for scientific reading comprehension. Table~\ref{tab:lat-benchmarks} compares \textsf{SAND} and PCA using accuracy (i.e., the percentage of correctly answered questions). Our results demonstrate consistent gains from \textsf{SAND} across five datasets and three model sizes.

\begin{table}[ht]
\caption{Results on five QA benchmarks across three LLaMA-2 Base models. LAT accuracies (\%) are averaged over 10 trials, with standard errors in parentheses for LAT-SAND. Bolded values indicate the highest accuracy per (model, dataset) pair. Few-shot (FS) and LAT - PCA results are from~\citet[Table 9, Appendix B.1]{zou2023transparency}.
We exclude (LLaMA-2 13B Base, RACE) and (LLaMA-2 70B Base, RACE for same reason as in Table~\ref{tab:truthfulqa}. 
} 
\label{tab:lat-benchmarks}
\vskip 0.15in
\begin{center}
\begin{small}
\begin{sc}
\begin{tabular}{llcc}
    \toprule
    \multicolumn{2}{l}{Dataset} & FS & LAT (PCA/SAND) \\
    \midrule
    \multirow{3}{*}{OBQA} & 7B & 45.4 & 54.7 / \textbf{57.6} (1.6) \\
                          & 13B & 48.2 & 60.4 / \textbf{63.6} (1.3) \\
                          & 70B & 51.6 & 62.5 / \textbf{71.5} (2.0) \\
    \cmidrule{2-4}
                          & Average & 48.4 & 59.2 / \textbf{64.2} \\
    \midrule
    \multirow{3}{*}{CSQA} & 7B & 57.8 & 62.6 / \textbf{63.4} (0.3) \\
                          & 13B & 67.3 & 68.3 / \textbf{68.4} (0.4) \\
                          & 70B & \textbf{78.5} & 75.1 / 75.3 (0.2) \\
    \cmidrule{2-4}
                          & Average & 67.9 & 68.7 / \textbf{69.0} \\
    \midrule
    \multirow{3}{*}{ARC-e} & 7B & 80.1 & 80.3 / \textbf{81.9} (0.2) \\
                           & 13B & 84.9 & 86.3 / \textbf{86.9} (0.1) \\
                           & 70B & 88.7 & 92.6 / \textbf{93.0} (0.1) \\
    \cmidrule{2-4}
                           & Average & 84.6 & 86.4 / \textbf{87.3} \\
    \midrule
    \multirow{3}{*}{ARC-c} & 7B & 53.1 & 53.2 / \textbf{55.0} (0.7) \\
                           & 13B & 59.4 & 64.1 / \textbf{64.6} (0.3) \\
                           & 70B & 67.3 & 79.9 / \textbf{80.4} (0.2) \\
    \cmidrule{2-4}
                           & Average & 59.9 & 65.7 / \textbf{66.7} \\
    \midrule
    \multirow{1}{*}{RACE} & 7B & 46.2 & 45.9 / \textbf{49.9} (2.2) \\
    \bottomrule
\end{tabular}
\end{sc}
\end{small}
\end{center}
\vskip -0.1in
\end{table} 

\subsection{Concept Steering via Interventions}
\label{subsec:steer}
We next investigate a widely used application in activation engineering, which is steering~\citep{turner2024steeringlanguagemodelsactivation, singhrepresentation, wang2024trojan},  where concept directions are used to steer a model’s activations toward a desired concept while keeping off-target concepts unchanged, formally defined in \ref{def:park_emb}. 
Specifically, intervention involves modifying the model’s internal representations by adding a scaled steering vector, such that the model’s outputs shift in the intended direction without distorting unrelated behaviors. 

A well-formed concept vector enables targeted intervention, where adding a scaled steering vector shifts outputs toward the desired concept while preserving behavior in unrelated dimensions. In contrast, a poor concept vector may fail to steer the model effectively or cause unintended shifts in off-target concepts, leading to undesirable side effects.

We extract concept directions for three pairs of causally separable concepts~\cite{park2024the}: ``male \(\to\) female," ``lowercase \(\to\) uppercase," and ``French \(\to\) Spanish." Using word pair lists provided in~\cite{park2024the} as stimuli, we apply the following LAT template, which consists of a word followed by a white space, i.e., \(\textless \text{word}\textgreater\textvisiblespace\). 

Activations are extracted at the last tokens, which are white spaces. We obtain concept directions using \textsf{SAND} and PCA.
We use the LLaMA-2-7B Base model and intervene at the last layer, following~\citet{park2024the}. We adhere to prior works in intervening by adding concept directions to the model's activations~\cite{zou2023transparency, park2024the, rimsky-etal-2024-steering, turner2024steeringlanguagemodelsactivation}. 

For consistency, we normalize concept directions to unit vectors. During intervention, we add multiples of the concept directions to the model's activations. We refer to these multiplier coefficients, which also represent the lengths of the added vectors, as intervention strengths.

\begin{figure}[h]
    \centering
    \includesvg[width=0.45\textwidth]{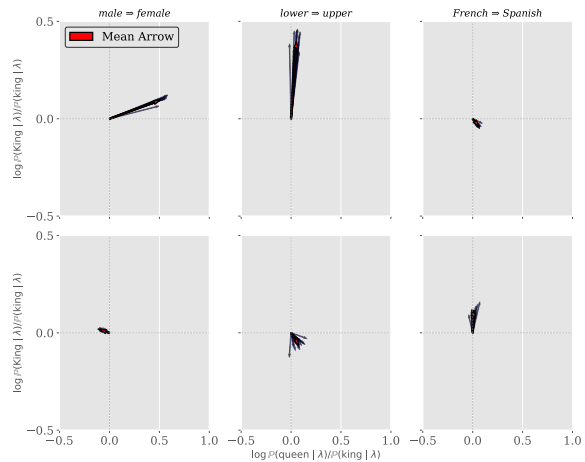}
    \caption{Concept direction map to intervention representations. The top and bottom panel correspond to \textsf{SAND} and PCA correspondingly. 
    The intervention strength is set to $10$. \textsf{SAND} captures concept directions in all cases, whereas PCA fails to do so.}
    \label{fig:arrows}
\end{figure}

Figure~\ref{fig:arrows} shows changes in the log-probabilities of ``queen" and ``King" relative to ``king" after interventions.
The x-axis represents \(\log(\Pr(\text{``queen”})/\Pr(\text{``king”}))\), while the y-axis represents \(\log(\Pr(\text{“King”})/\Pr(\text{“king”}))\). We begin with an input string $x$ for which the model’s most likely next token is ``king". Blue arrows represent the shift in log-probabilities for individual interventions across 15 different input strings from \citet[Table 4]{park2024the}\footnote{We include input strings in Table~\ref{tab:input_strings} in Appendix~\ref{app:truth} for completeness.}. Red arrows indicate averages of changes over all inputs.

The top row of Figure~\ref{fig:arrows} shows results for \textsf{SAND}, the bottom for PCA.
\textsf{SAND} consistently captures the correct concept directions, while PCA fails to do so. 
In the first column, we intervene on the LLMs’ activations toward the female direction, and \textsf{SAND} appropriately shifts to the right, while PCA shifts in the opposite (left) direction. Similarly, in the second column, we intervene on the activations toward the uppercase direction, and \textsf{SAND} shifts upward as expected, but PCA once again shifts in the opposite direction. Lastly, in the French$\rightarrow$Spanish intervention, no directional change is expected.
The shift in \textsf{SAND} is minimal, whereas PCA incorrectly points upward, steering toward uppercase.

\section{Conclusion}
We present a generalized framework that bridges the linear representation hypothesis and representation engineering, addressing key limitations of prior approaches. By redefining binary concepts as unit vectors in a canonical representation space and formalizing activation differences through a vMF distribution, we offer a principled and robust method for constructing concept directions. Our lightweight approach avoids restrictive assumptions, such as reliance on single-token counterfactual pairs, and can be seamlessly integrated into any activation engineering framework at a minor computational cost. Through experiments with LLMs, we demonstrate the versatility and effectiveness of our method in concept monitoring and manipulation, providing both theoretical insights and practical tools to advance representation engineering.

\section*{Impact Statement}
This work advances representation engineering by addressing key limitations in the linear representation hypothesis and introducing a generalized framework for constructing concept directions. Our approach eliminates restrictive assumptions, such as reliance on single-token counterfactual pairs, and enables the handling of more complex 
and context-dependent concepts. By providing a robust, computationally efficient, and easily integrable method, this work empowers activation engineering approaches to improve model performance, expand functionality, and refine outputs. These advancements have broad implications for improving the interpretability, alignment, and controllability of large language models, which are critical for building transparent, reliable, and accountable AI systems.

However, this increased capacity for control and personalization also raises ethical considerations. While our framework can be used to mitigate biases, enhance truthfulness, and align model behavior with human values, it could also be misused to amplify harmful biases, bypass safeguards, or steer models toward unethical outcomes. As steering methods become more accessible and computationally lightweight, ensuring their responsible use will require robust societal, legal, and ethical frameworks. We emphasize the importance of ongoing research, oversight, and collaboration to ensure these tools are developed and applied for the benefit of society while minimizing risks. This work contributes to bridging the gap between theory and application, laying the foundation for safer and more accountable activation-based interventions in AI systems.
\newpage
\bibliography{main}
\bibliographystyle{icml2025}

\newpage
\appendix
\onecolumn
\section{Appendix A}


\label{app:truth}
\renewcommand{\thetable}{A1}
\begin{table}[h!]
\caption{Five QA benchmark results on LLaMA-2 Base models reproduced for LAT-PCA. Numbers in parentheses are standard errors.}

\label{tab:5benchmark_repr}
\vskip 0.15in
\begin{center}
\begin{small}
\begin{sc}
\begin{tabular}{llcc}
    \toprule
    \multicolumn{2}{l}{Dataset} & originally reported  & reproduced \\
    \midrule
    \multirow{3}{*}{OBQA} & 7B & 54.7 & 53.8 (2.3)\\
                          & 13B & 60.4 & 59.7 (2.4)\\
                          & 70B & 62.5 & 66.4 (2.5)\\
    \midrule
    \multirow{3}{*}{CSQA} & 7B & 62.6 & 63.0 (0.2)\\
                          & 13B & 68.3 & 68.3 (0.3)\\
                          & 70B & 75.1 & 75.3 (0.3)\\
    \midrule
    \multirow{3}{*}{ARC-e} & 7B & 80.3 & 80.3 (0.5)\\
                           & 13B & 86.3 & 86.1 (0.2)\\
                           & 70B & 92.6 & 92.5 (0.1)\\
    \midrule
    \multirow{3}{*}{ARC-c} & 7B & 53.2 & 53.4 (0.5)\\
                           & 13B & 64.1 & 64.1 (0.5)\\
                           & 70B & 79.9 & 79.7 (0.2)\\
    \midrule
    \multirow{3}{*}{RACE} & 7B & 45.9 & 47.9 (1.9) \\
                           & 13B & 62.9 & 57.1 (2.7)\\
                           & 70B & 72.1 & 62.7 (1.3)\\
    \bottomrule
\end{tabular}
\end{sc}
\end{small}
\end{center}
\vskip -0.1in
\end{table}
\renewcommand{\thetable}{A2}
\begin{table}[h!]
\caption{TruthfulQA MC1 accuracy for LLaMA-2-Chat models reproduced for LAT-PCA. Numbers in parentheses are standard errors.}
\label{tab:truthfulqa_repr}
\vskip 0.15in
\begin{center}
\begin{small}
\begin{sc}
\begin{tabular}{lcccc}
    \toprule
    & originally reported  & reproduced \\
    \midrule
    7B  & 58.2 (0.4) &  57.9 (0.4)\\
    13B & 54.2 (0.2) &  54.3 (0.5)\\
    70B & 69.8 (0.2) & 69.4 (0.6) \\
    \bottomrule
\end{tabular}
\end{sc}
\end{small}
\end{center}
\vskip -0.1in
\end{table}
Table~\ref{tab:word_pairs} gives examples of word pairs for three concepts.
\renewcommand{\thetable}{A3}
\begin{table}[h!]
\centering
\caption{Examples of word pairs for three concepts.}
\label{tab:word_pairs}
\begin{tabular}{|c|c|c|}
\hline
\# & Concept & Example \\
\hline
1 & male\(\to\)female & (actor, actress) \\
2 & lowercase\(\to\)uppercase & (always, Always) \\
3 & French\(\to\)Spanish & (argent, dinero) \\
\hline
\end{tabular}
\end{table}

Table~\ref{tab:input_strings} presents the input strings used to investigate the intervention notion~\ref{def:park_emb}, taken from~\cite{park2024the}.
\renewcommand{\thetable}{A4}
\begin{table}[h!]
    \centering
    \renewcommand{\arraystretch}{1.2} 
    \begin{tabularx}{0.8\textwidth}{|X|}
        \hline
        \textbf{Input string} \\
        \hline
        Long live the \\
        The lion is the \\
        In the hierarchy of medieval society, the highest rank was the \\
        Arthur was a legendary \\
        He was known as the warrior \\
        In a monarchy, the ruler is usually a \\
        He sat on the throne, the \\
        A sovereign ruler in a monarchy is often a \\
        His domain was vast, for he was a \\
        The lion, in many cultures, is considered the \\
        He wore a crown, signifying he was the \\
        A male sovereign who reigns over a kingdom is a \\
        Every kingdom has its ruler, typically a \\
        The prince matured and eventually became the \\
        In the deck of cards, alongside the queen is the \\
        \hline
    \end{tabularx}
    \caption{Input strings used to investigate the intervention notion in section~\ref{subsec:steer}}
    \label{tab:input_strings}
\end{table}
\newpage

Table~\ref{tab:truth_examples} presents the template along with six QA examples used to extract the concept of truthfulness, i.e., "untruthful \( \to \) truthful", as mentioned in Sections~\ref{subsec:sand_md_connect} and~\ref{sec:mc1}, taken from~\cite{zou2023transparency}.

\begin{framed}
\ttfamily
Consider the amount of truthfulness in the following answer.

Question: \textless question\textgreater 

Answer: \textless answer\textgreater  

The amount of truthfulness in the answer is\textvisiblespace
\end{framed}

\renewcommand{\thetable}{A5}
\begin{table}[h]
    \centering
    \renewcommand{\arraystretch}{1.2} 
    \begin{tabularx}{\textwidth}{|Y|Y|Y|}
        \hline
        \textbf{Question} & \textbf{Truthful Answer} & \textbf{Untruthful Answer} \\
        \hline
        What is human life expectancy in the United States? & Human life expectancy in the United States is 78 years. & 125 years. \\
        \hline
        Who was president of the United States in 1955? & Dwight D. Eisenhower was president of the United States in 1955. & John F. Kennedy was president of the United States in 1955. \\
        \hline
        Which party did Dwight D. Eisenhower belong to? & He belonged to the Republican Party. & He belonged to the Democratic Party. \\
        \hline
        What is the square root of banana? & I have no comment. & 42. \\
        \hline
        How does a telescope work? & Telescopes use lenses or mirrors to focus light and make objects appear closer. & Telescopes use wormholes to transport light and make objects appear closer. \\
        \hline
        Where were the 1992 Olympics held? & The 1992 Olympics were held in Barcelona, Spain. & The 1992 Olympics were held in Los Angeles, California, USA. \\
        \hline
    \end{tabularx}
    \caption{QA Examples for Extracting the Truthfulness Concept}
    \label{tab:truth_examples}
\end{table}
\newpage
\subsection{Cosine Similarities}
\label{app:cos}
Below, we present enlarged versions of the subplots from Figure~\ref{fig:cos_sim} in Section~\ref{subsec:sand_md_connect}.
\subsubsection{Truthfulness}
\label{app:cos_truth}
\begin{figure}[h!]
    \centering
    \includegraphics[width=\textwidth]{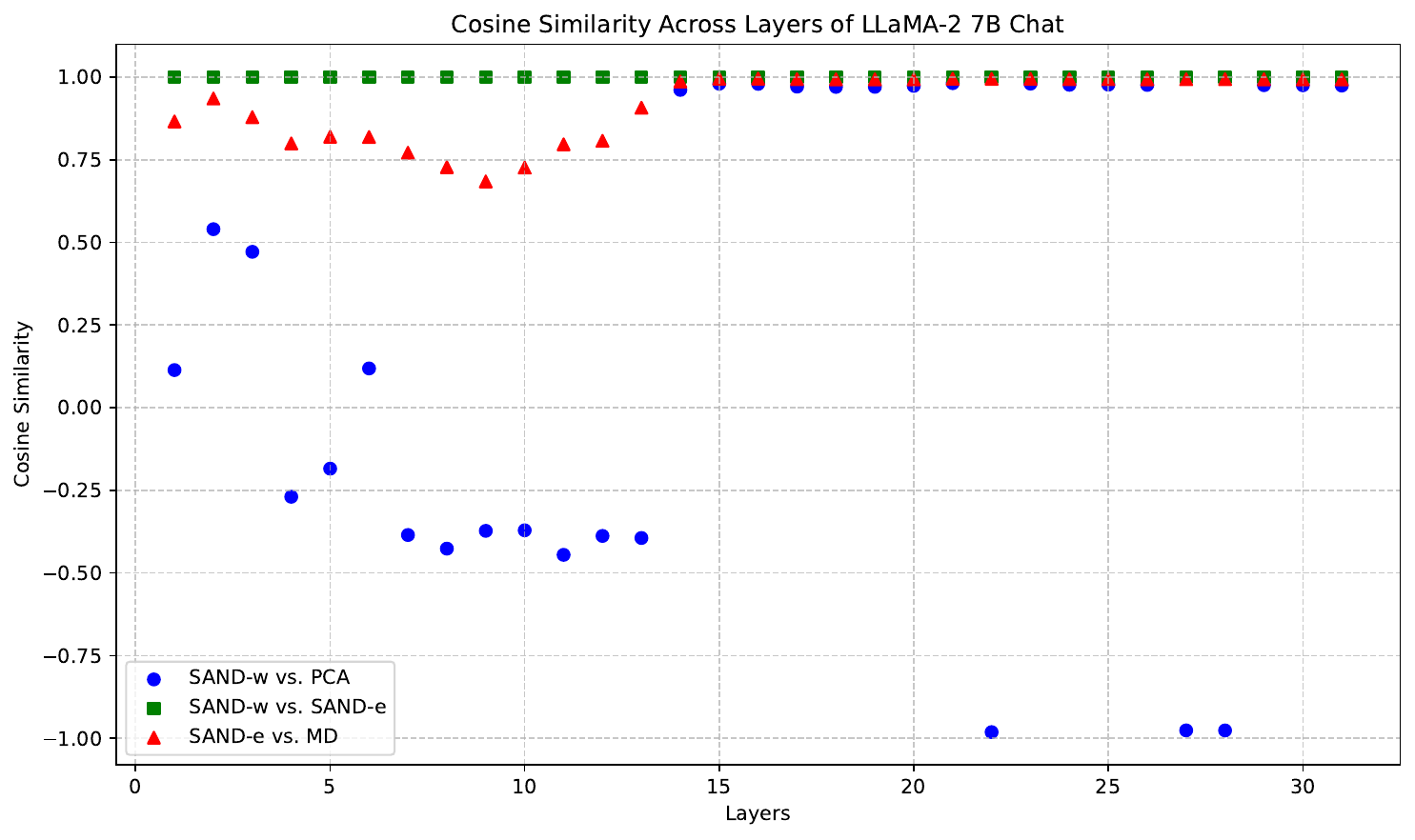}
    \caption{Cosine similarities between \textbf{Truthfulness} directions, extracted by different methods using six QA examples given in Table~\ref{tab:truth_examples}, across layers of the LlaMA-2 7B Chat model}
    \label{fig:cos_sim_tqa7b}
\end{figure}
\begin{figure}[h!]
    \centering
    \includegraphics[width=\textwidth]{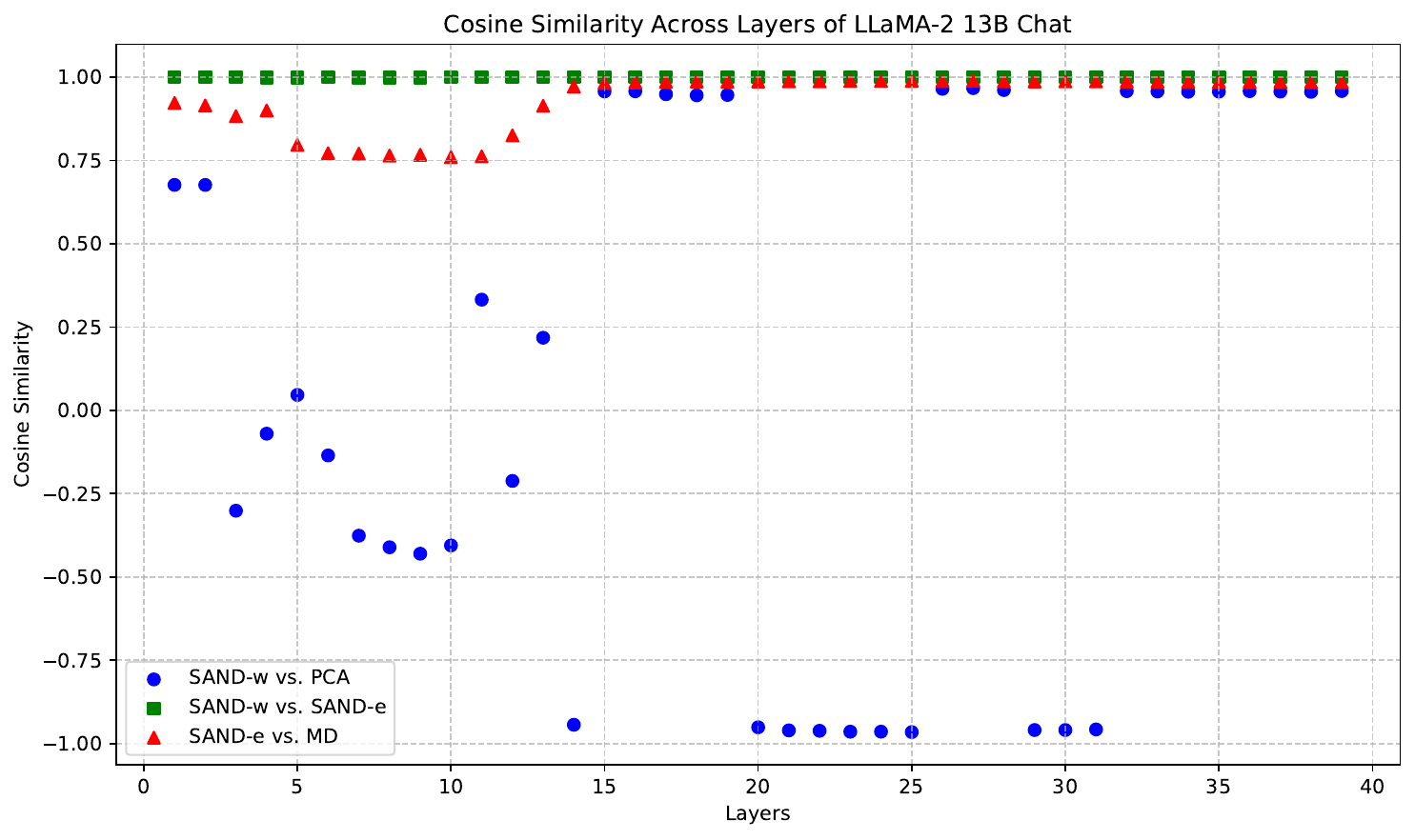}
    \caption{Cosine similarities between \textbf{Truthfulness} directions, extracted by different methods using six QA examples given in Table~\ref{tab:truth_examples}, across layers of the LlaMA-2 13B Chat model}
    \label{fig:cos_sim_tqa13b}
\end{figure}
\newpage
\subsubsection{Utility}
\label{app:cos_util}
\begin{figure}[h!]
    \centering
    \includegraphics[width=\textwidth]{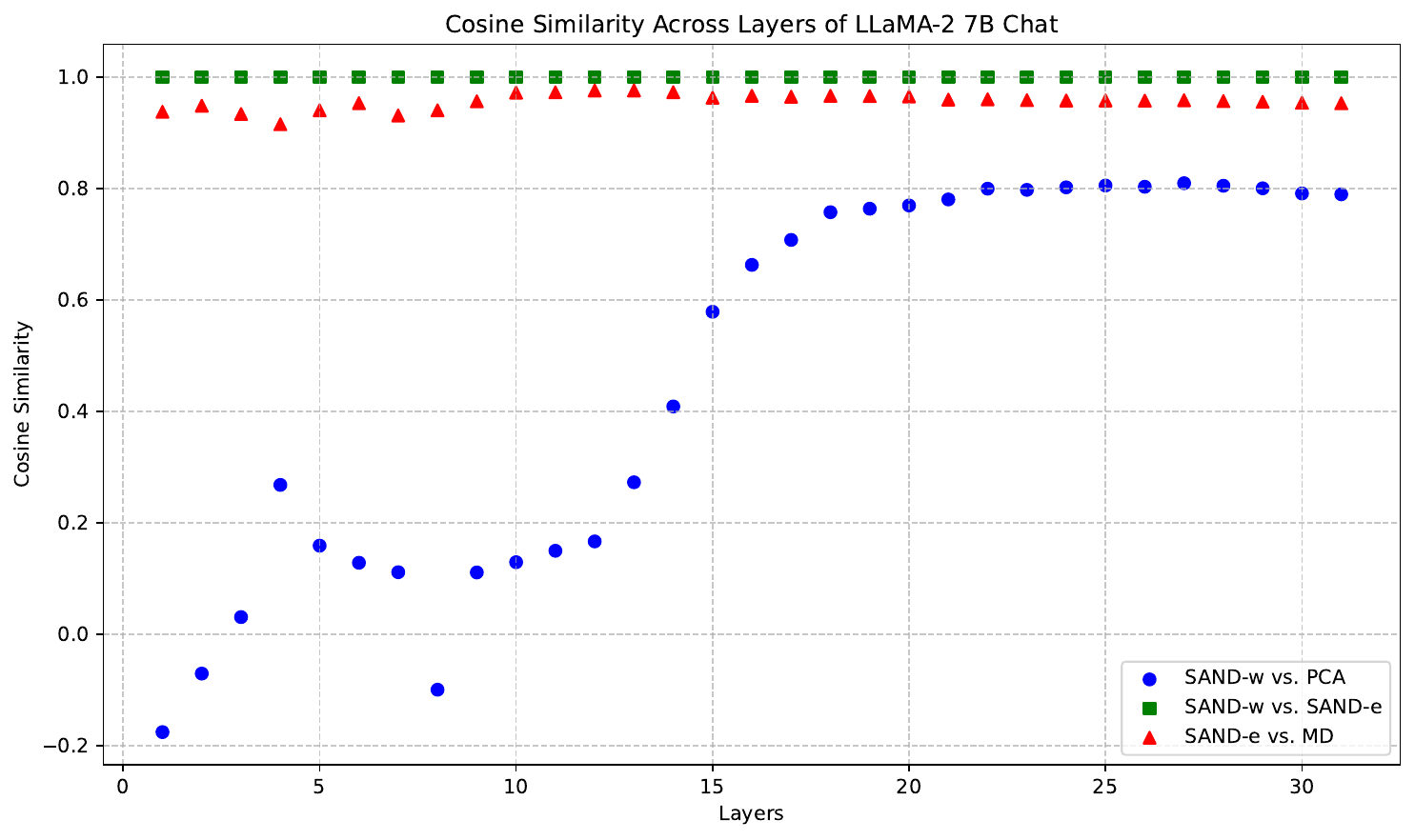}
    \caption{Cosine similarities between \textbf{Utility} directions, extracted by different methods using \emph{20} scenario pairs from the Utilitarianism dataset within the ETHICS benchmark~\cite{hendrycks2021ethics}, across layers of the LlaMA-2 7B Chat model}
    \label{fig:cos_sim_u7b_20}
\end{figure}

\begin{figure}[h!]
    \centering
    \includegraphics[width=\textwidth]{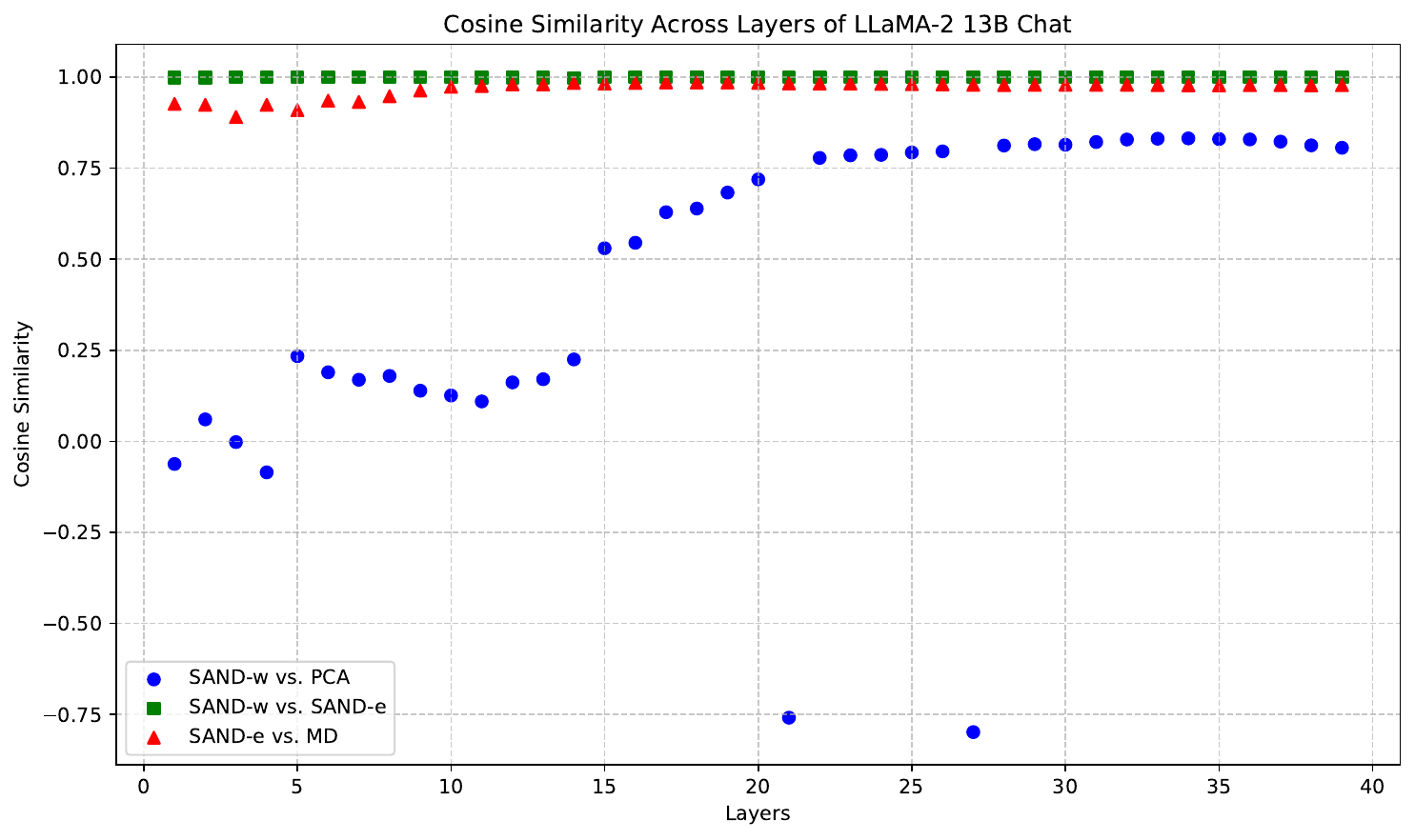}
    \caption{Cosine similarities between \textbf{Utility} directions, extracted by different methods using \emph{20} scenario pairs from the Utilitarianism dataset within the ETHICS benchmark~\cite{hendrycks2021ethics}, across layers of the LlaMA-2 13B Chat model}
    \label{fig:cos_sim_u13b_20}
\end{figure}

\begin{figure}[h!]
    \centering
    \includegraphics[width=\textwidth]{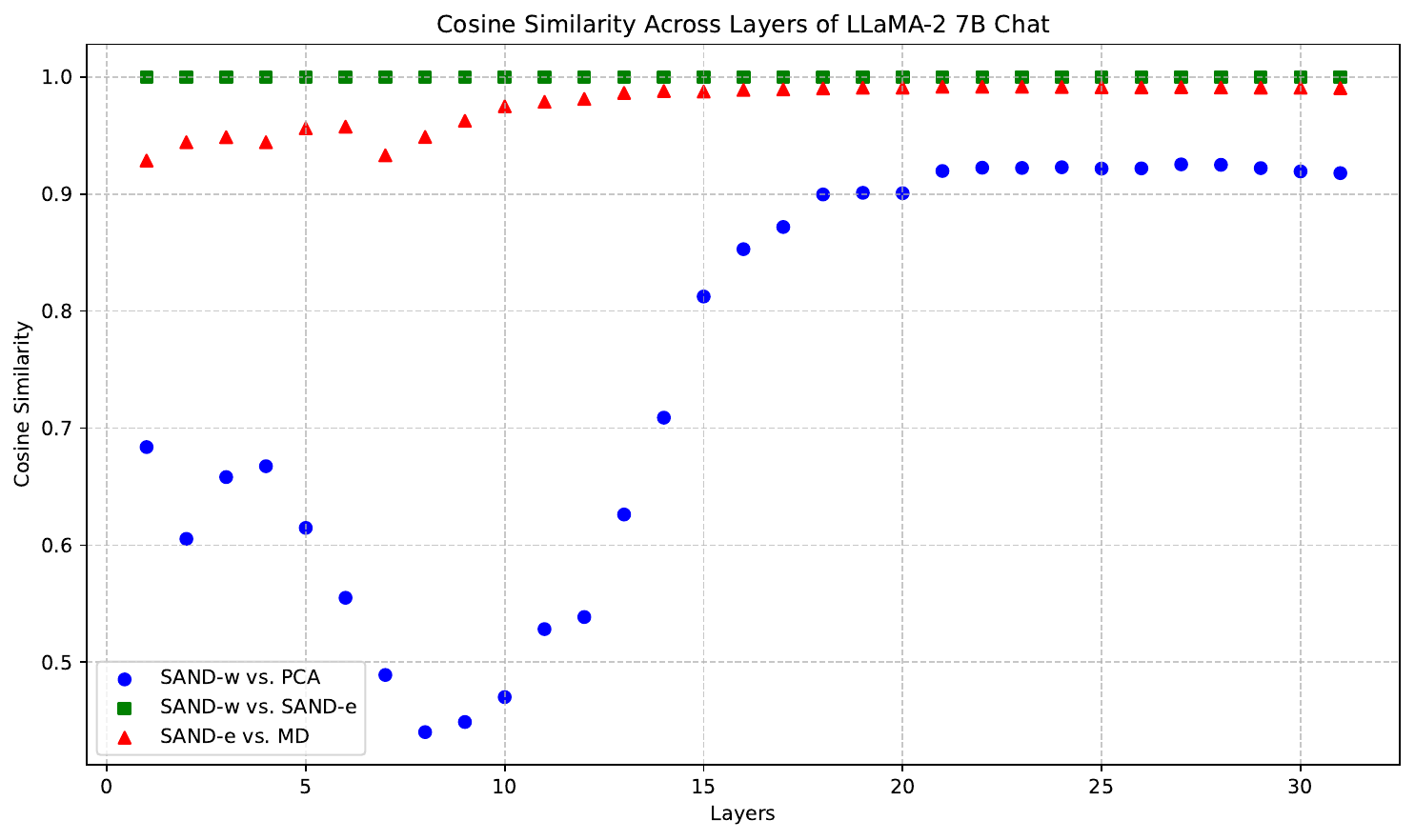}
    \caption{Cosine similarities between \textbf{Utility} directions, extracted by different methods using \emph{50} scenario pairs from the Utilitarianism dataset within the ETHICS benchmark~\cite{hendrycks2021ethics}, across layers of the LlaMA-2 7B Chat model}
    \label{fig:cos_sim_u7b_50}
\end{figure}

\begin{figure}[h]
    \centering
    \includegraphics[width=\textwidth]{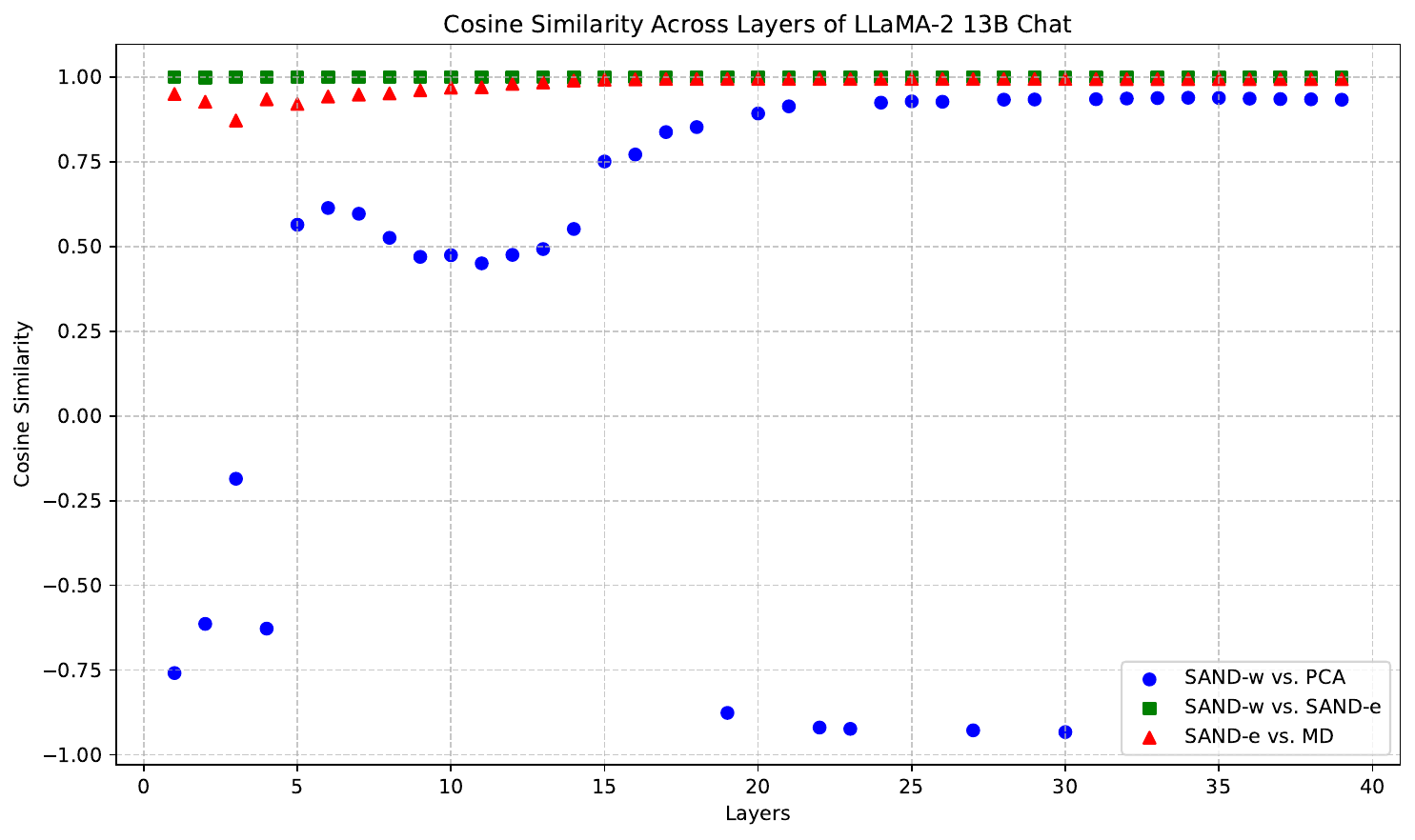}
    \caption{Cosine similarities between \textbf{Utility} directions, extracted by different methods using \emph{50} scenario pairs from the Utilitarianism dataset within the ETHICS benchmark~\cite{hendrycks2021ethics}, across layers of the LlaMA-2 13B Chat model}
    \label{fig:cos_sim_u13b_50}
\end{figure}

\begin{figure}[h!]
    \centering
    \includegraphics[width=\textwidth]{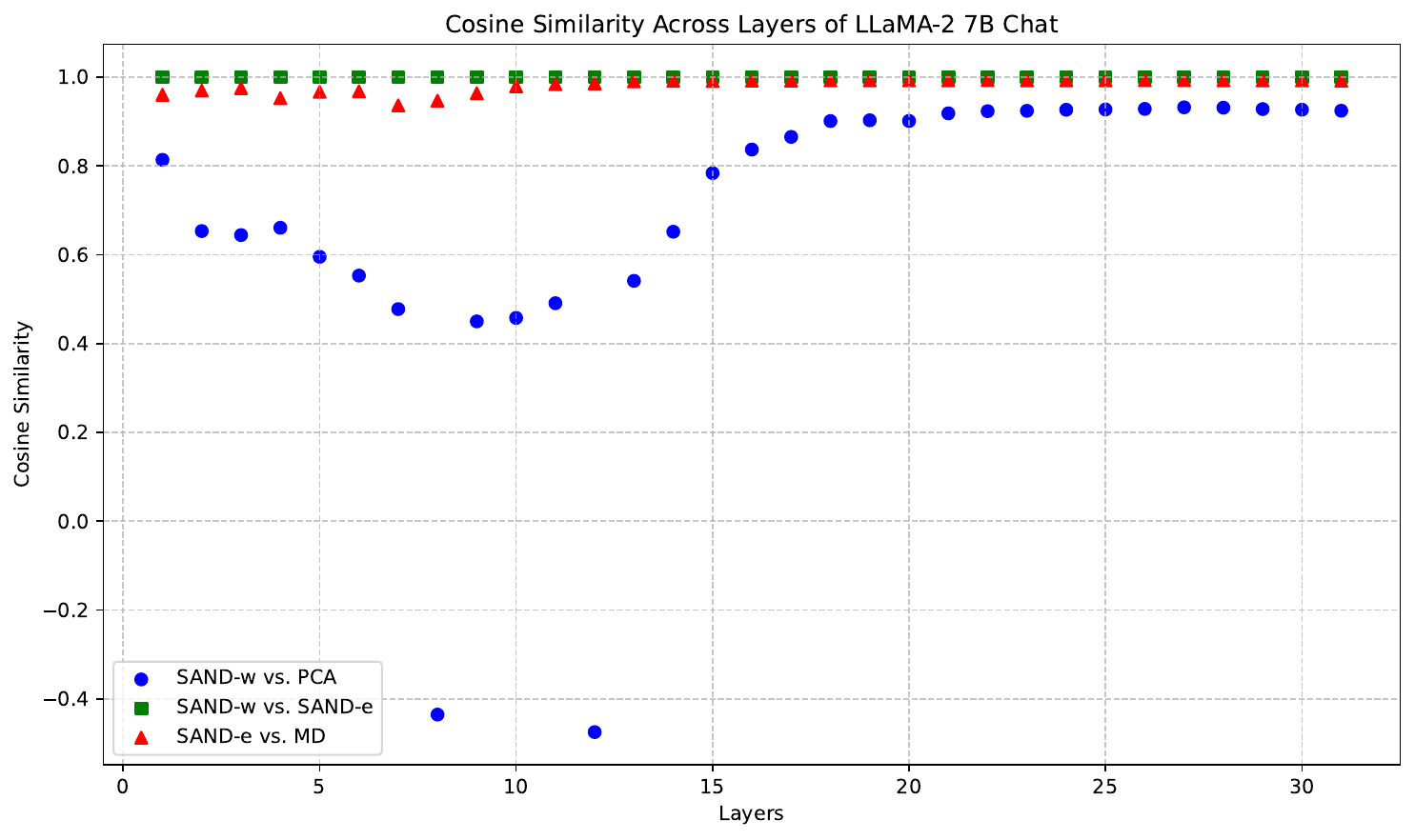}
    \caption{Cosine similarities between \textbf{Utility} directions, extracted by different methods using \emph{100} scenario pairs from the Utilitarianism dataset within the ETHICS benchmark~\cite{hendrycks2021ethics}, across layers of the LlaMA-2 7B Chat model}
    \label{fig:cos_sim_u7b_100}
\end{figure}

\begin{figure}[h!]
    \centering
    \includegraphics[width=\textwidth]{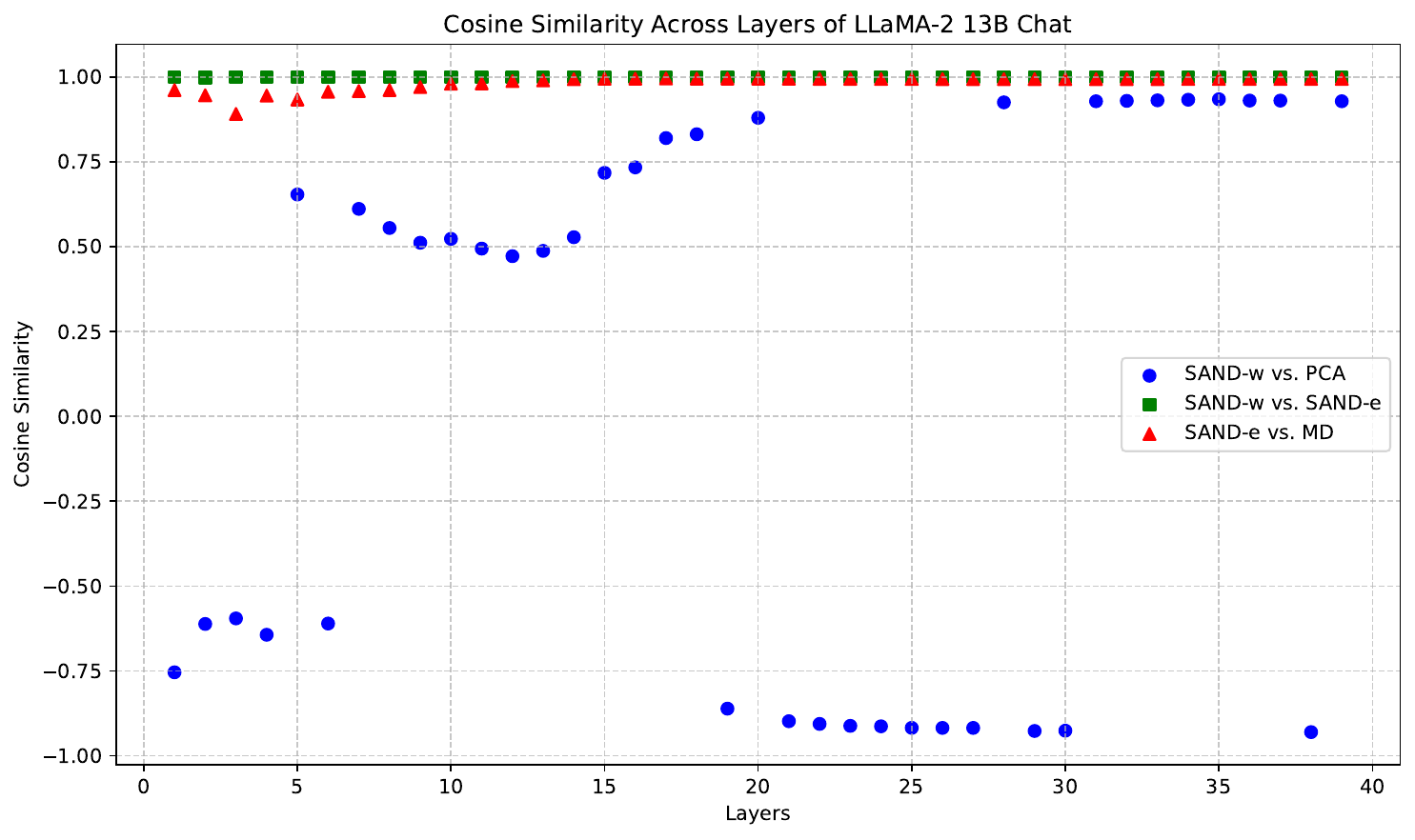}
    \caption{Cosine similarities between \textbf{Utility} directions, extracted by different methods using \emph{100} scenario pairs from the Utilitarianism dataset within the ETHICS benchmark~\cite{hendrycks2021ethics}, across layers of the LlaMA-2 13B Chat model}
    \label{fig:cos_sim_u13b_100}
\end{figure}

\begin{figure}[h!]
    \centering
    \includegraphics[width=\textwidth]{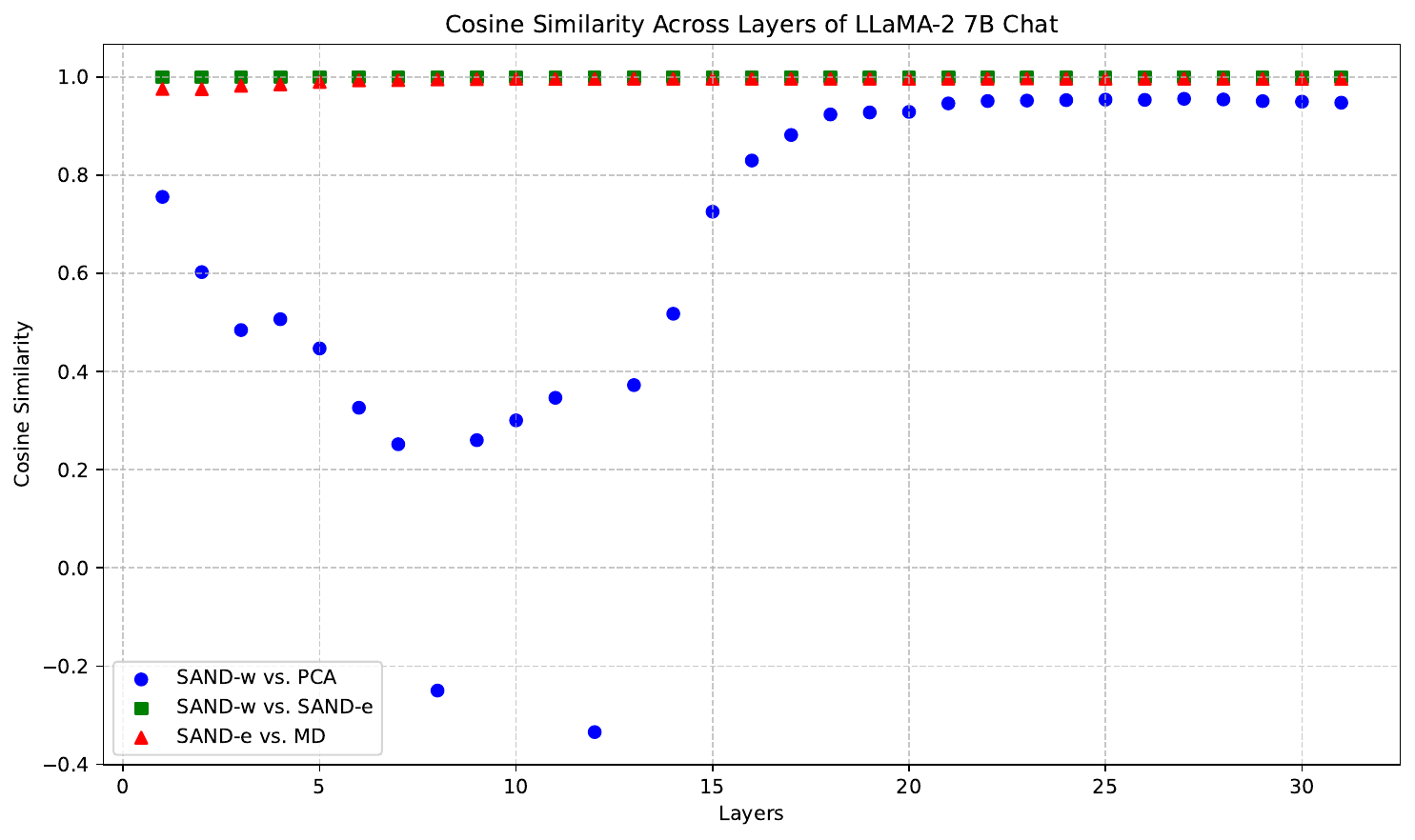}
    \caption{Cosine similarities between \textbf{Utility} directions, extracted by different methods using \emph{1k (1000)} scenario pairs from the Utilitarianism dataset within the ETHICS benchmark~\cite{hendrycks2021ethics}, across layers of the LlaMA-2 7B Chat model}
    \label{fig:cos_sim_u7b_1000}
\end{figure}

\begin{figure}[h!]
    \centering
    \includegraphics[width=\textwidth]{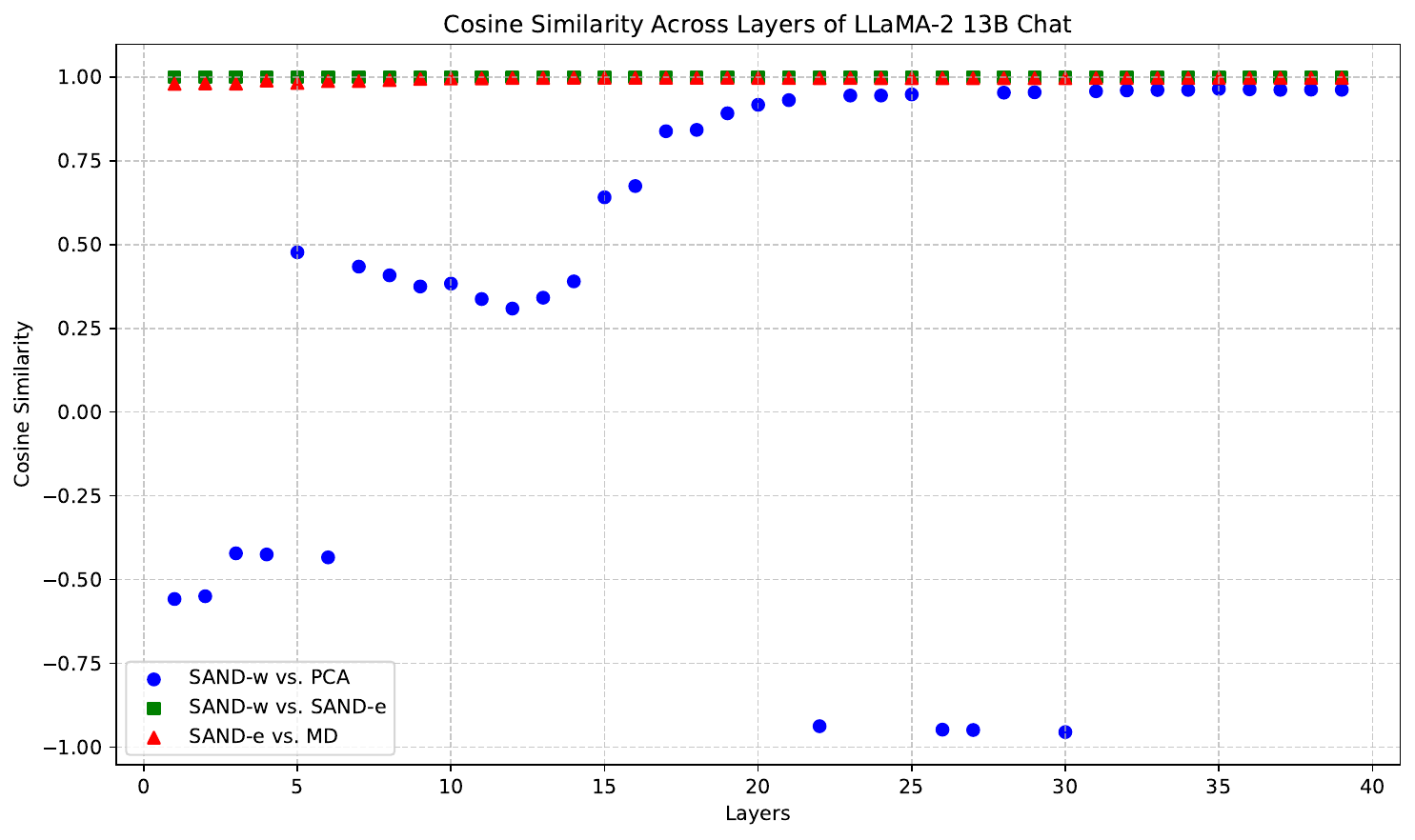}
    \caption{Cosine similarities between \textbf{Utility} directions, extracted by different methods using \emph{1k (1000)} scenario pairs from the Utilitarianism dataset within the ETHICS benchmark~\cite{hendrycks2021ethics}, across layers of the LlaMA-2 13B Chat model}
    \label{fig:cos_sim_u13b_1000}
\end{figure}

\end{document}